\documentclass[conference]{IEEEtran}
\IEEEoverridecommandlockouts
\usepackage{cite}
\usepackage{amsmath,amssymb,amsfonts}
\usepackage{graphicx}
\usepackage{textcomp}
\usepackage{xcolor}
\usepackage{algpseudocode}
    \makeatletter
\newcommand{\linebreakand}{%
  \end{@IEEEauthorhalign}
  \hfill\mbox{}\par
  \mbox{}\hfill\begin{@IEEEauthorhalign}
}
\makeatother
\begin{document}

\title{Exploring the Efficacy of Federated-Continual Learning Nodes with Attention-Based Classifier for Robust Web Phishing Detection: An Empirical Investigation

}

\author{\IEEEauthorblockN{Jesher Joshua M}
\IEEEauthorblockA{\textit{School of Computer Science and Engineering} \\
\textit{Vellore Institute of Technology}\\
Chennai, India \\
jesherjoshua.m2021@vitstudent.ac.in}
\and
\IEEEauthorblockN{Adhithya R}
\IEEEauthorblockA{\textit{School of Computer Science and Engineering} \\
\textit{Vellore Institute of Technology}\\
Chennai, India \\
adhithya.r2021@vitstudent.ac.in}
\linebreakand
%\and <------------------------ REMOVED
\IEEEauthorblockN
{Sree Dananjay S}
\IEEEauthorblockA{\textit{School of Computer Science and Engineering} \\
\textit{Vellore Institute of Technology}\\
Chennai, India \\
sreedananjay.s2021@vitstudent.ac.in}
\and
\IEEEauthorblockN{M Revathi}
\IEEEauthorblockA{\textit{School of Computer Science and Engineering} \\
\textit{Vellore Institute of Technology}\\
Chennai, India \\
m.revathi@vit.ac.in}
}

\maketitle

\begin{abstract}
Web phishing poses a dynamic threat, requiring detection systems to quickly adapt to the latest tactics. Traditional approaches of accumulating data and periodically retraining models are outpaced. We propose a novel paradigm combining federated learning and continual learning, enabling distributed nodes to continually update models on streams of new phishing data, without accumulating data. These locally adapted models are then aggregated at a central server via federated learning. To enhance detection, we introduce a custom attention-based classifier model with residual connections, tailored for web phishing, leveraging attention mechanisms to capture intricate phishing patterns. We evaluate our hybrid learning paradigm across continual learning strategies (cumulative, replay, MIR, LwF) and model architectures through an empirical investigation. Our main contributions are: (1) a new hybrid federated-continual learning paradigm for robust web phishing detection, and (2) a novel attention + residual connections based model explicitly designed for this task, attaining 0.93 accuracy, 0.90 precision, 0.96 recall and 0.93 f1-score with the LwF strategy, outperforming traditional approaches in detecting emerging phishing threats while retaining past knowledge.
\end{abstract}


\section{Introduction}
Phishing attacks continue to pose a significant threat to online security, exploiting human vulnerabilities through deceptive tactics to gain unauthorized access to sensitive information or systems. These attacks often involve the creation of fraudulent websites that mimic legitimate ones, luring unsuspecting users into divulging personal data or credentials. As phishing techniques become increasingly sophisticated, the need for robust and adaptable detection mechanisms becomes paramount.

Traditional approaches to phishing website detection have relied on static blacklists or heuristic-based methods, which often lag behind the rapidly evolving nature of these attacks. Machine learning (ML) techniques have emerged as a promising solution, offering the capability to learn and identify complex patterns indicative of phishing websites. However, most existing ML-based approaches suffer from several limitations, such as the inability to adapt to new phishing strategies, reliance on outdated datasets, and performance degradation over time.

To address these challenges, we propose a novel hybrid learning paradigm that combines the strengths of federated learning and continual learning for robust web phishing detection. Federated learning enables distributed learning nodes to collaboratively train a shared model without centralizing data, preserving privacy and data sovereignty. Continual learning, on the other hand, allows these nodes to continually adapt their models to the most recent phishing data streams, ensuring timely detection of emerging threats.
Our approach leverages the power of attention mechanisms to develop a tailored attention-based classifier model explicitly designed for web phishing detection. By capturing intricate patterns and contextual cues indicative of phishing websites, this model enhances the accuracy and robustness of the detection process. Furthermore, we incorporate adaptive feature selection mechanisms to identify the most relevant features dynamically, further improving the model's performance and interpretability.
The integration of federated and continual learning paradigms addresses the dynamic nature of phishing attacks, enabling distributed nodes to continually update their models based on the latest phishing data streams. This approach eliminates the need for data accumulation and periodic retraining, ensuring that the detection system remains up-to-date and effective against evolving threats.

Through an extensive empirical investigation, we evaluate the efficacy of our proposed hybrid learning paradigm across various continual learning strategies, model architectures, and datasets. We compare our approach with traditional ML techniques and state-of-the-art methods, demonstrating its superior performance in detecting the latest phishing threats while preserving knowledge from past data distributions.

The main contributions of this research are twofold: (1) a novel hybrid learning paradigm that combines federated and continual learning for robust web phishing detection in dynamic environments, and (2) a tailored attention-based classifier model designed explicitly for web phishing detection, leveraging attention mechanisms and adaptive feature selection.

By addressing the limitations of existing approaches and offering a comprehensive solution for robust and adaptable phishing detection, our work contributes significantly to the ongoing efforts in mitigating this persistent cyber threat, ultimately enhancing online security and protecting users from falling victim to these deceptive attacks.

\section{Related works}
Phishing website detection remains a critical challenge in cybersecurity, prompting extensive research efforts to develop effective countermeasures. Numerous studies have focused on employing machine learning (ML) techniques for phishing detection. Zamir et al. \cite{1} proposed a stacking model approach combining multiple ML algorithms, while Salloum et al. \cite{5} compared various classifiers, including XGBoost, Random Forest, and ANN. Similarly, Alnemari and Alshammari \cite{8} evaluated ANNs, SVMs, DTs, and RF for detecting phishing domains. However, these traditional ML techniques often struggle to adapt to the dynamic nature of phishing attacks and may suffer from performance deterioration over time.

To address this limitation, our work introduces a novel hybrid learning paradigm that combines federated learning and continual learning. By enabling distributed learning nodes to continuously adapt their models to the most recent phishing data streams, our approach ensures robust phishing detection in dynamic environments without the need for data accumulation. This approach aligns with the work of Ejaz et al. \cite{11}, who advocated for continual learning as a sustainable solution for phishing attack detection over time.

Several studies have explored deep learning architectures for phishing detection. Dutta \cite{3} employed a recurrent neural network, while Maci et al. \cite{7} utilized the double deep Q-Network (DDQN) for unbalanced web phishing classification. Nagy et al. \cite{4} developed a parallel computing ML model using CNN and LSTM for real-time detection. In contrast, our work introduces a novel attention-based classifier model tailored specifically for web phishing detection, leveraging attention mechanisms to capture intricate patterns and contextual cues indicative of phishing websites.

Feature engineering and selection have been crucial aspects in phishing detection research. Sánchez-Paniagua et al. \cite{6} highlighted the limitations of existing datasets and proposed a new dataset containing essential elements like login samples. Das Guptta et al. \cite{12} considered URL and hyperlink-based hybrid features, while Abdul Samad et al. \cite{15} investigated the impact of feature selection and hyperparameter optimization. Our approach complements these efforts by incorporating attention-based feature selection mechanisms into our novel classifier model, adaptively identifying the most relevant features for accurate phishing detection.

Several studies have explored hybrid and ensemble approaches. Jovanovic et al. \cite{16} introduced a hybrid two-level framework for feature selection and XGBoost tuning, while Sakhare et al. \cite{14} combined multiple algorithms, including XGBoost, LightGBM, and Graph Neural Networks. Prabakaran et al. \cite{17} integrated Variational Autoencoders (VAE) with deep neural networks. Our hybrid learning paradigm leverages the strengths of federated and continual learning, presenting a unique and effective solution for robust phishing detection in dynamic environments.

Researchers have also investigated alternative techniques, such as Bozkir et al. \cite{18} proposing the GramBeddings model using n-gram embeddings, Ozcan et al. \cite{19} combining character embedding-based and NLP features in a hybrid DNN-LSTM model, and Kumar et al. \cite{20} employing swarm intelligence optimization for network design. While these approaches contribute valuable insights, our work focuses on a comprehensive solution that addresses the key challenges of adaptability, feature selection, and model robustness in phishing detection.

In summary, our proposed approach stands out by introducing a novel hybrid learning paradigm that combines the benefits of federated and continual learning, enabling distributed nodes to continually adapt their models to the most recent phishing data streams. Additionally, our work introduces a tailored attention-based classifier model designed explicitly for web phishing detection, leveraging attention mechanisms and adaptive feature selection to capture intricate phishing patterns effectively. By addressing the limitations of traditional ML techniques and offering a comprehensive solution for robust and adaptable phishing detection, our approach contributes significantly to the ongoing efforts in mitigating this persistent cyber threat.

\section{About the dataset}
The dataset used in this study is the result of merging two publicly available datasets: the "Web page phishing detection" dataset \cite{21} and the "Phishing Websites Dataset" \cite{22}. A subset of the most relevant features was selectively included in the merged dataset to avoid redundancy and focus on the shared characteristics between the two original datasets. The resulting dataset provides a comprehensive view of the shared features while maintaining a streamlined and focused set of attributes.

Initially, the raw dataset exhibited class imbalance, with a disproportionate number of samples in normal class compared to the phishing. To address this issue, undersampling was performed, where the majority class was downsampled to match the size of the minority class, resulting in a balanced dataset.

\begin{figure}[htbp]
\centering
\includegraphics[width=1.0\linewidth]{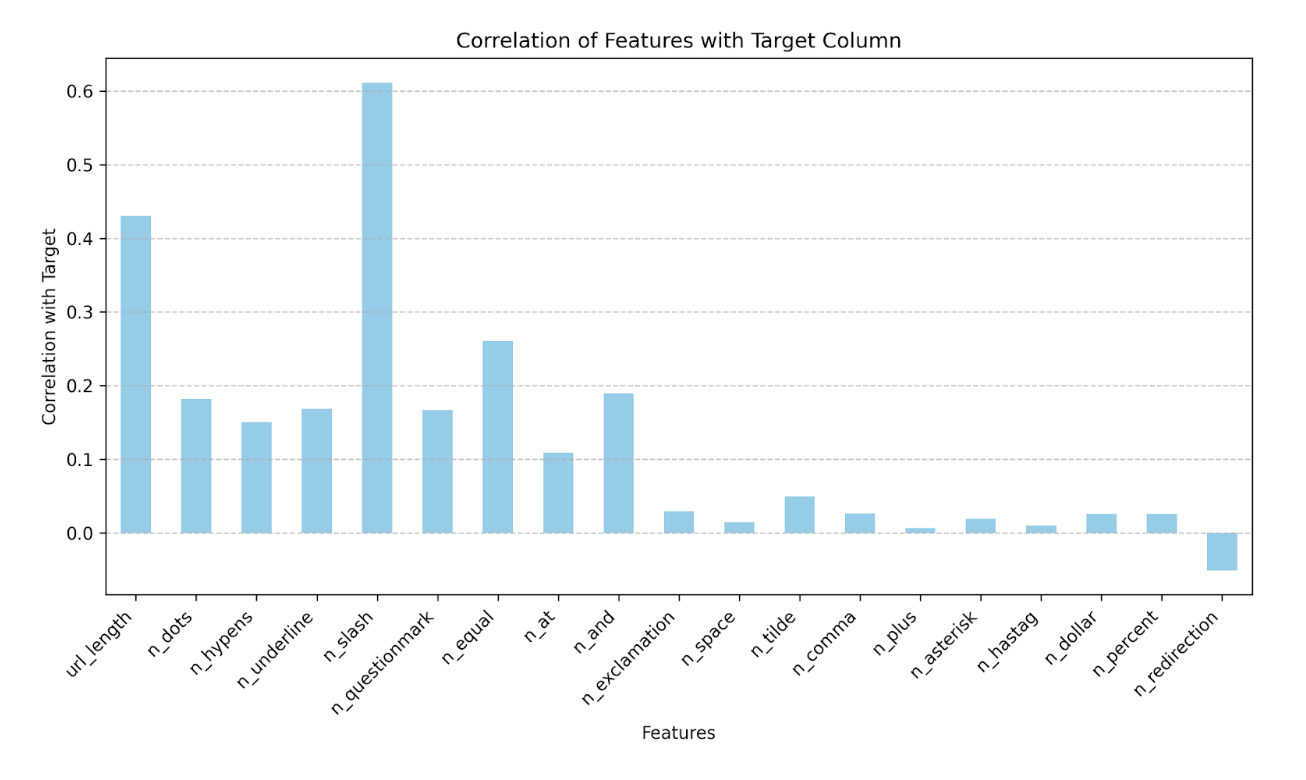} % Adjust the width as needed
\caption{Features correlation with target.}
\label{fig_data}
\end{figure}

The dataset with each row representing a website and each column representing a feature. The last column contains the binary label indicating whether the website is phishing (1) or legitimate (0). The dataset consists of 20 features related to the characteristics of the URL. These features include the length of the URL, the count of various special characters (such as dots, hyphens, underscores, slashes, question marks, equal signs, at symbols, ampersands, exclamation marks, spaces, tildes, commas, plus signs, asterisks, hashtags, dollar signs, and percent signs), and the count of redirections within the URL. Fig. \ref{fig_data}, illustrates the correlation of the feature columns with the target.

\section{Proposed Solution}
\begin{figure}[htbp]
\centering
\includegraphics[width=1.0\linewidth]{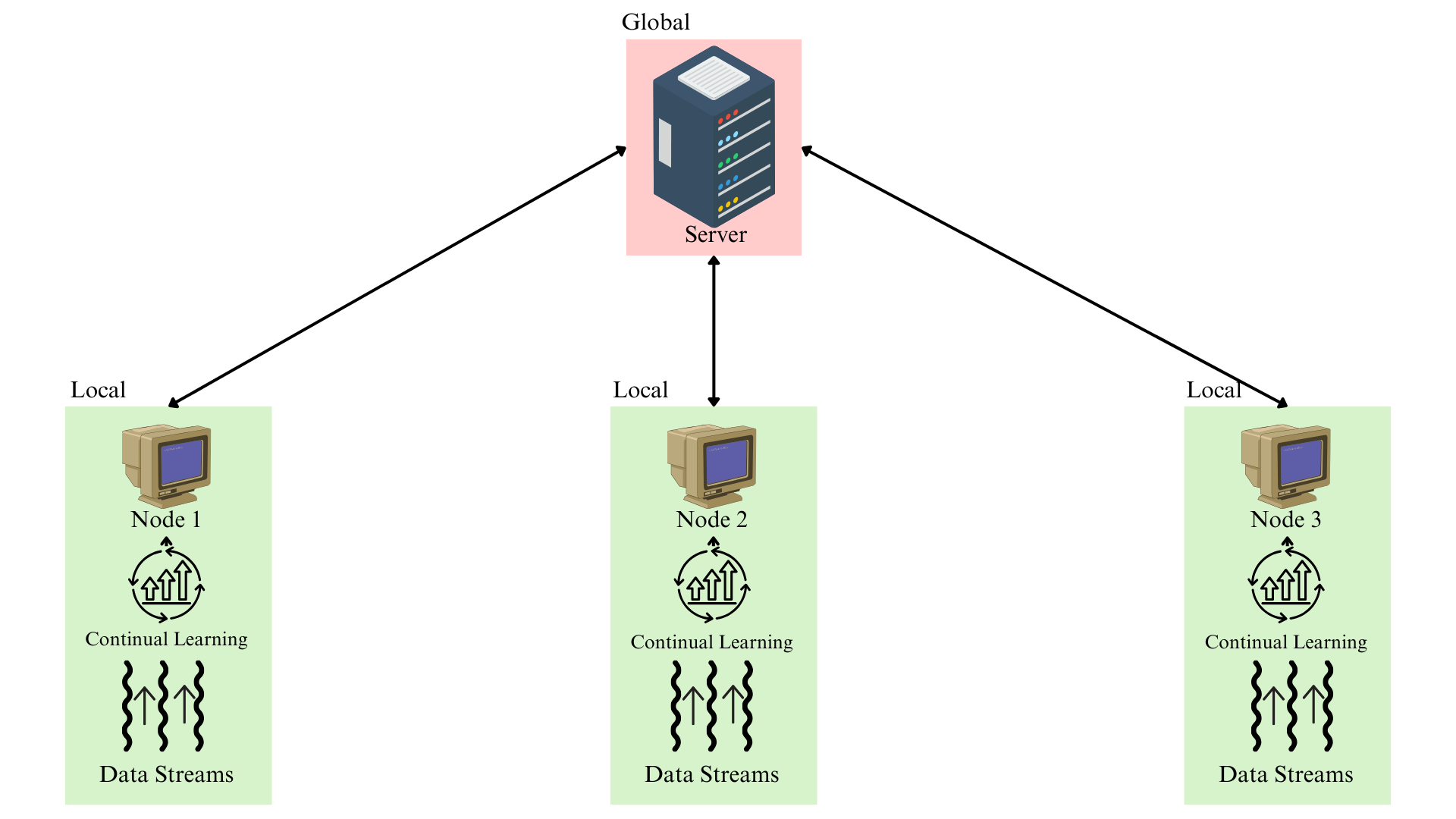} % Adjust the width as needed
\caption{Schematic of proposed solution.}
\label{fig_proposed}
\end{figure}

To address the limitations of traditional machine learning approaches in detecting phishing websites and adapt to the dynamic nature of these attacks, we propose a novel hybrid learning paradigm that seamlessly integrates federated learning and continual learning. This integrative approach leverages the strengths of both learning paradigms, enabling distributed nodes to continually update their models based on the most recent phishing data streams while preserving user privacy and data sovereignty. Fig. \ref{fig_proposed}, provides an overview of our proposed solution.
\subsection{Federated Learning for Collaborative Model Training}
Federated learning is a decentralized approach to machine learning that enables multiple parties to collaboratively train a shared model without exchanging raw data. In our proposed solution, distributed learning nodes (e.g., client devices, organizations, or edge nodes) train local models on their respective phishing data streams. These local models are then aggregated at a central server through a secure federated averaging protocol, which updates the global model without exposing individual data points.

The federated learning framework offers several advantages in the context of phishing website detection. First, it preserves user privacy by keeping sensitive data on local devices, eliminating the need for centralized data storage. Second, it enables collaborative learning from diverse data sources, enhancing the model's ability to capture a wide range of phishing patterns and strategies. Third, it fosters data sovereignty, allowing organizations or individuals to retain control over their data while benefiting from the collective knowledge of the federated network.

\subsection{Continual Learning for Adapting to Evolving Threats
}
While federated learning facilitates collaborative model training, the dynamic nature of phishing attacks demands a continuous learning approach to ensure the detection system remains effective against emerging threats. Continual learning, also known as lifelong learning, enables machine learning models to adapt to new data distributions without catastrophically forgetting previously learned knowledge.

In our proposed solution, the distributed learning nodes employ continual learning strategies to continually update their local models based on the most recent phishing data streams. We investigate various continual learning techniques, including cumulative learning, replay-based methods (e.g., experience replay), regularization-based approaches (e.g., MIR, LwF), and others, to identify the most effective strategy for phishing website detection.

\subsubsection{Naive Continual Learning}
Naive continual learning, also known as incremental learning, is a straightforward approach where a model is trained continuously on new data without considering the previously learned knowledge. While this strategy is simple to implement, it often leads to catastrophic forgetting, where the model loses performance on previously learned tasks as it focuses solely on the most recent data. Naive continual learning lacks mechanisms to mitigate forgetting and adapt to new tasks without compromising performance on older ones.

\subsubsection{Cumulative Continual Learning\cite{24}}
Cumulative continual learning aims to address the issue of catastrophic forgetting by preserving previously learned knowledge while incorporating new information. This strategy involves updating the model parameters in a way that minimizes interference with existing knowledge when learning new tasks. Techniques such as parameter regularization, rehearsal, and task-specific constraints are employed to ensure that the model retains its performance on previous tasks while adapting to new ones. Cumulative continual learning is effective in maintaining a balance between old and new knowledge, thereby enabling the model to learn sequentially without significant performance degradation.

\subsubsection{Replay Continual Learning}
Replay continual learning mitigates catastrophic forgetting by periodically replaying samples from previous tasks during training. These replayed samples serve as a form of rehearsal, allowing the model to reinforce its understanding of earlier tasks while learning new ones. Replay can be implemented using various techniques, such as storing a buffer of past experiences or generating synthetic data resembling previous tasks. By exposing the model to a diverse range of past experiences, replay continual learning helps to preserve and consolidate previously acquired knowledge, thus improving overall performance across multiple tasks over time.

\subsubsection{Learning without Forgetting (LwF)\cite{23}}
Learning without Forgetting (LwF) is a continual learning strategy that aims to retain performance on previous tasks while learning new ones without explicit rehearsal or replay. LwF achieves this by leveraging the knowledge distillation technique, where the model is trained to mimic its own predictions from previous iterations when learning new tasks. By distilling the knowledge from the old model into the new one, LwF encourages the model to preserve its understanding of previous tasks while adapting to new ones. This approach effectively reduces the risk of catastrophic forgetting by ensuring that the model's knowledge is retained and transferred across tasks throughout the learning process.

\subsubsection{Maximally Interfered Replay (MIR)\cite{25}}
Maximally Interfered Replay (MIR) is a continual learning strategy that draws inspiration from human memory mechanisms to mitigate forgetting. MIR maintains a memory buffer containing representations of past experiences, which are selectively sampled and replayed alongside new data during training. By integrating past experiences into the learning process, MIR enables the model to continually update its knowledge while retaining information about previous tasks. The selective sampling mechanism ensures that the memory buffer remains relevant to the current learning task, optimizing the balance between old and new information. MIR's ability to leverage past experiences for continual learning makes it effective in preserving and consolidating knowledge across multiple tasks over time.

By integrating continual learning into the federated learning framework, our approach ensures that the global model remains up-to-date and adaptable to evolving phishing strategies. As new phishing data becomes available at the local nodes, their continually updated models are aggregated through federated averaging, continuously refining the global model's knowledge and detection capabilities.

\subsection{Attention-Based Classifier Model}
\begin{figure*}[htbp]
\centering
\includegraphics[width=1.0\linewidth]{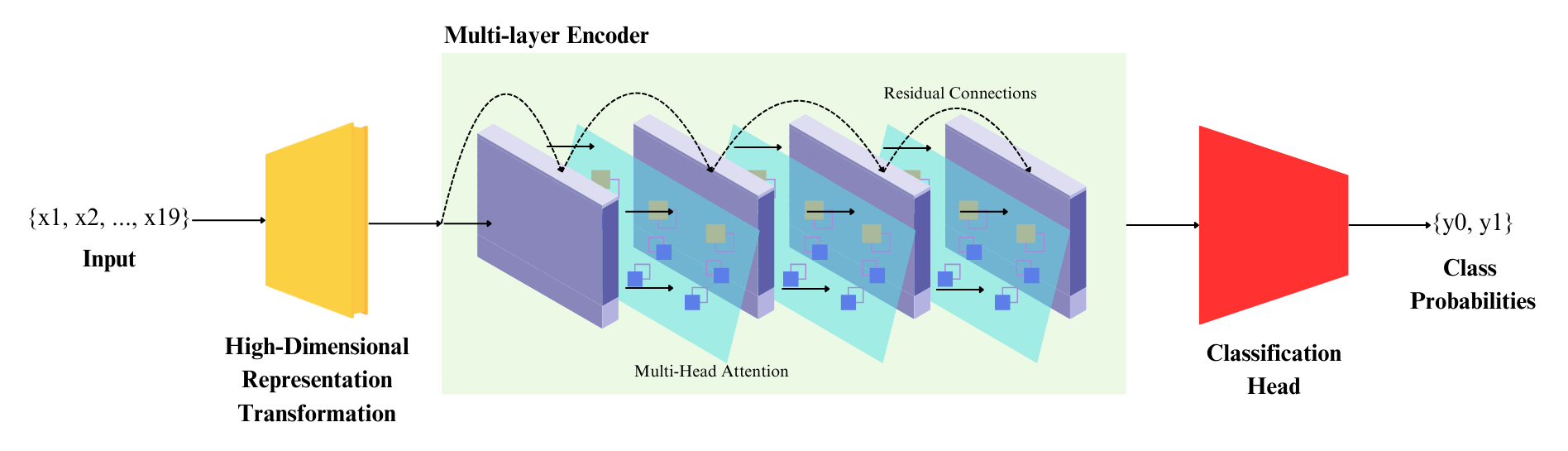} % Adjust the width as needed
\caption{Custom neural network  architecture.}
\label{fig_model}
\end{figure*}

At the core of our proposed solution is a novel attention-based classifier model tailored explicitly for web phishing detection. In Fig. \ref{fig_model}, our proposed attention-based classifier model is designed to effectively capture the intricate patterns and contextual cues indicative of phishing websites, leveraging the power of attention mechanisms. The input data is first projected into a higher-dimensional space through a fully connected linear layer. This initial encoding step aims to capture a rich representation of the input data. 

The encoded representation is then passed through a multi-layer encoder, which consists of a sequence of linear transformations interleaved with residual connections and multi-head attention mechanisms. Specifically, the encoder is implemented as a module list of linear layers, where each layer performs a linear transformation by projecting the input into the hidden dimension. The transformed input is then combined with the original input through a residual connection, allowing for better gradient flow and mitigating the vanishing gradient problem. Subsequently, the residual output is passed through a multi-head attention mechanism, which computes attention weights over the input sequence, enabling the model to focus on the most relevant features or patterns.

The multi-head attention mechanism is a crucial component of our model, as it allows the model to attend to different representations of the input data by projecting the queries, keys, and values into multiple subspaces and computing attention weights separately in each subspace. This mechanism is particularly beneficial for phishing website detection, as it can help the model identify and prioritize the discriminative features and patterns that distinguish phishing websites from legitimate ones. After the multi-layer encoder, the encoded representation is passed through a dropout layer to regularize the model and prevent overfitting.

Finally, the encoded representation is mapped to the output classes through a fully connected linear layer, producing logits that represent the model's predictions for the binary classification task (phishing or legitimate website). During the forward pass, the input data is first encoded and then propagated through the multi-layer encoder. At each layer, the residual connection and multi-head attention mechanism allow the model to capture and attend to the most relevant features and patterns. The encoded representation is then processed by the dropout layer and finally passed through the classification layer to obtain the output logits.
\subsection{Proposed Algorithm}
Our proposed algorithm Fig. \ref{euclid}, seamlessly integrates federated learning, continual learning, and the attention-based classifier model to provide a robust and adaptable solution for phishing website detection. 
\begin{figure}

\begin{algorithmic}[1]
\Procedure{Hybrid-learning}{Initial global model $\mathbf{w}_0$, number of nodes $N$, learning rates $\{\eta_k\}_{k=1}^N$}
\State Initialize local models $\mathbf{w}_k^0 = \mathbf{w}_0$ for all nodes $k = 1, \ldots, N$
\State Initialize cumulative weight $\mathbf{w}_{\text{cum}} = \mathbf{0}$
\State Initialize total sample size $n_{\text{total}} = 0$
\For{$t = 1, 2, \ldots$}
    \State Wait for a node $k$ to send its updated model $\mathbf{w}_k^t$ and local sample size $n_k^t$
    \State $n_{\text{total}} \gets n_{\text{total}} + n_k^t$
    \State $\mathbf{w}_{\text{cum}} \gets \mathbf{w}_{\text{cum}} + n_k^t \mathbf{w}_k^t$
    \State $\mathbf{w}_0 \gets \frac{1}{n_{\text{total}}} \mathbf{w}_{\text{cum}}$
    \State Broadcast the updated global model $\mathbf{w}_0$ to all nodes
\EndFor
\EndProcedure
\end{algorithmic}
\caption{Proposed hybrid-learning algorithm}
\label{euclid}
\end{figure}

The proposed algorithm leverages dynamic averaging(7,8) to aggregate the local model updates from the participating nodes, ensuring that the global model remains up-to-date and adaptable to emerging phishing threats. The attention-based classifier model is optimized during the local model updates, capturing the most relevant features and patterns for phishing detection through the attention mechanism.

By combining federated learning, continual learning, and the attention-based classifier model, our algorithm facilitates collaborative training, continuous adaptation to evolving threats, and effective pattern recognition through attention mechanisms. This approach ensures that the global model remains up-to-date and adaptable to emerging phishing strategies, while preserving user privacy and data sovereignty.

\section{Result Analysis}

To evaluate the efficacy of our proposed hybrid learning paradigm and attention-based classifier model, we conduct an extensive empirical investigation. We compare our approach to pre-existing neural network architectures and investigate the best continual learning strategy for this scenario.

Our experimental setup involves multiple datasets, to assess the performance and adaptability of our solution under various scenarios. We evaluate our approach across different continual learning strategies and model architectures, aiming to identify the optimal setup for robust and accurate phishing detection.

\begin{table}[htbp]
\caption{Experimental Setup Hyper-Parameters}
\begin{center}
\begin{tabular}{|l|l|}
\hline
\textbf{Hyper-Parameter} & \textbf{Value} \\ \hline
Federated Learning Rounds & 20 \\ \hline
Local Nodes & 3 \\ \hline
Global Server & 1 \\ \hline
Epochs in Local Node & 10 \\ \hline
Learning Rate & 0.001 \\ \hline
Data Streams & 4 \\ \hline
Batch Size & 16 \\ \hline
\end{tabular}
\label{tab:hyperparameters}
\end{center}
\end{table}
The evaluation metrics considered in our analysis include accuracy, precision, recall and F1-score across various continual learning strategies, namely Naive, Replay, Cumulative, Learning without Forgetting (LwF), and Maximally Interfered Replay (MIR). Each strategy is compared against traditional machine learning models including Simple Multilayer Perceptron (MLP), Deep MLP, and Simple Recurrent Neural Network (RNN).

In Table \ref{tab:naive_results} the Naive continual learning strategy, our attention-based classifier model achieved an accuracy of 0.70, outperforming Simple MLP, Deep MLP, and Simple RNN models with accuracies of 0.52, 0.50, and 0.64, respectively. The precision, recall, and F1-score for our model were 0.75, 0.60, and 0.67, indicating its effectiveness in capturing phishing website patterns compared to baseline models.
\begin{table}[htbp]
\caption{Results under Naive Strategy}
\begin{center}
\begin{tabular}{|l|l|l|l|l|}
\hline
\textbf{Model} & \textbf{Accuracy} & \textbf{Precision} & \textbf{Recall} & \textbf{F1-Score} \\ \hline
Simple MLP & 0.52 & 0.25 & 0.50 & 0.33 \\ \hline
Deep MLP & 0.50 & 0.43 & 0.75 & 0.55 \\ \hline
Simple RNN & 0.64 & 0.86 & 0.50 & 0.62 \\ \hline
Ours & 0.70 & 0.75 & 0.60 & 0.67 \\ \hline
\end{tabular}
\label{tab:naive_results}
\end{center}
\end{table}

In Table \ref{tab:replay_results} under the Replay continual learning strategy, our model demonstrated further improvement with an accuracy of 0.73, surpassing Simple MLP, Deep MLP, and Simple RNN models with accuracies of 0.60, 0.65, and 0.67, respectively. Notably, our model achieved the highest precision and F1-score of 0.91 and 0.75, respectively, highlighting its superior ability in distinguishing phishing websites.

\begin{table}[htbp]
\caption{Results under Replay Strategy}
\begin{center}
\begin{tabular}{|l|l|l|l|l|}
\hline
\textbf{Model} & \textbf{Accuracy} & \textbf{Precision} & \textbf{Recall} & \textbf{F1-Score} \\ \hline
Simple MLP & 0.60 & 0.70 & 0.57 & 0.64 \\ \hline
Deep MLP & 0.65 & 0.78 & 0.60 & 0.67 \\ \hline
Simple RNN & 0.67 & 0.80 & 0.63 & 0.71 \\ \hline
Ours & 0.73 & 0.91 & 0.63 & 0.75 \\ \hline
\end{tabular}
\label{tab:replay_results}
\end{center}
\end{table}

In Table \ref{tab:cumulative_results} our attention-based classifier model excelled in the Cumulative continual learning scenario, achieving an accuracy of 0.86, substantially outperforming Simple MLP, Deep MLP, and Simple RNN models with accuracies of 0.72, 0.76, and 0.76, respectively. Additionally, our model exhibited superior precision, recall, and F1-score of 0.87, 0.84, and 0.84, underscoring its robustness and adaptability in cumulative learning settings.
\begin{table}[htbp]
\caption{Results under Cumulative Strategy}
\begin{center}
\begin{tabular}{|l|l|l|l|l|}
\hline
\textbf{Model} & \textbf{Accuracy} & \textbf{Precision} & \textbf{Recall} & \textbf{F1-Score} \\ \hline
Simple MLP & 0.72 & 0.73 & 0.75 & 0.73 \\ \hline
Deep MLP & 0.76 & 0.78 & 0.75 & 0.77 \\ \hline
Simple RNN & 0.76 & 0.77 & 0.77 & 0.77 \\ \hline
Ours & 0.86 & 0.87 & 0.84 & 0.84 \\ \hline
\end{tabular}
\label{tab:cumulative_results}
\end{center}
\end{table}

In Table \ref{tab:lwfs_results} the LwF continual learning approach, our model demonstrated remarkable performance, achieving an accuracy of 0.93, significantly outperforming Simple MLP, Deep MLP, and Simple RNN models with accuracies of 0.77, 0.73, and 0.88, respectively. With precision, recall, and F1-score values of 0.90, 0.96, and 0.93, our model exhibited superior capability in preserving previously learned knowledge while accommodating new information.

\begin{table}[htbp]
\caption{Results under LwF Strategy}
\begin{center}
\begin{tabular}{|l|l|l|l|l|}
\hline
\textbf{Model} & \textbf{Accuracy} & \textbf{Precision} & \textbf{Recall} & \textbf{F1-Score} \\ \hline
Simple MLP & 0.77 & 0.69 & 0.75 & 0.71 \\ \hline
Deep MLP & 0.73 & 0.67 & 0.71 & 0.69 \\ \hline
Simple RNN & 0.88 & 0.87 & 0.87 & 0.87 \\ \hline
Ours & 0.93 & 0.90 & 0.96 & 0.93 \\ \hline
\end{tabular}
\label{tab:lwfs_results}
\end{center}
\end{table}

In Table \ref{tab:mir_results} under the MIR continual learning strategy, our attention-based classifier model achieved an accuracy of 0.59, surpassing Simple MLP, Deep MLP, and Simple RNN models with accuracies of 0.53, 0.55, and 0.54, respectively. Despite the lower performance compared to other strategies, our model maintained competitive precision, recall, and F1-score values of 0.58 each, showcasing its potential in adapting to evolving data distributions.

\begin{table}[htbp]
\caption{Results under MIR Strategy}
\begin{center}
\begin{tabular}{|l|l|l|l|l|}
\hline
\textbf{Model} & \textbf{Accuracy} & \textbf{Precision} & \textbf{Recall} & \textbf{F1-Score} \\ \hline
Simple MLP & 0.53 & 0.56 & 0.53 & 0.53 \\ \hline
Deep MLP & 0.55 & 0.57 & 0.53 & 0.55 \\ \hline
Simple RNN & 0.54 & 0.56 & 0.53 & 0.55 \\ \hline
Ours & 0.59 & 0.58 & 0.58 & 0.58 \\ \hline
\end{tabular}
\label{tab:mir_results}
\end{center}
\end{table}

\begin{figure}[htbp]
\centering
\includegraphics[width=0.8\linewidth]{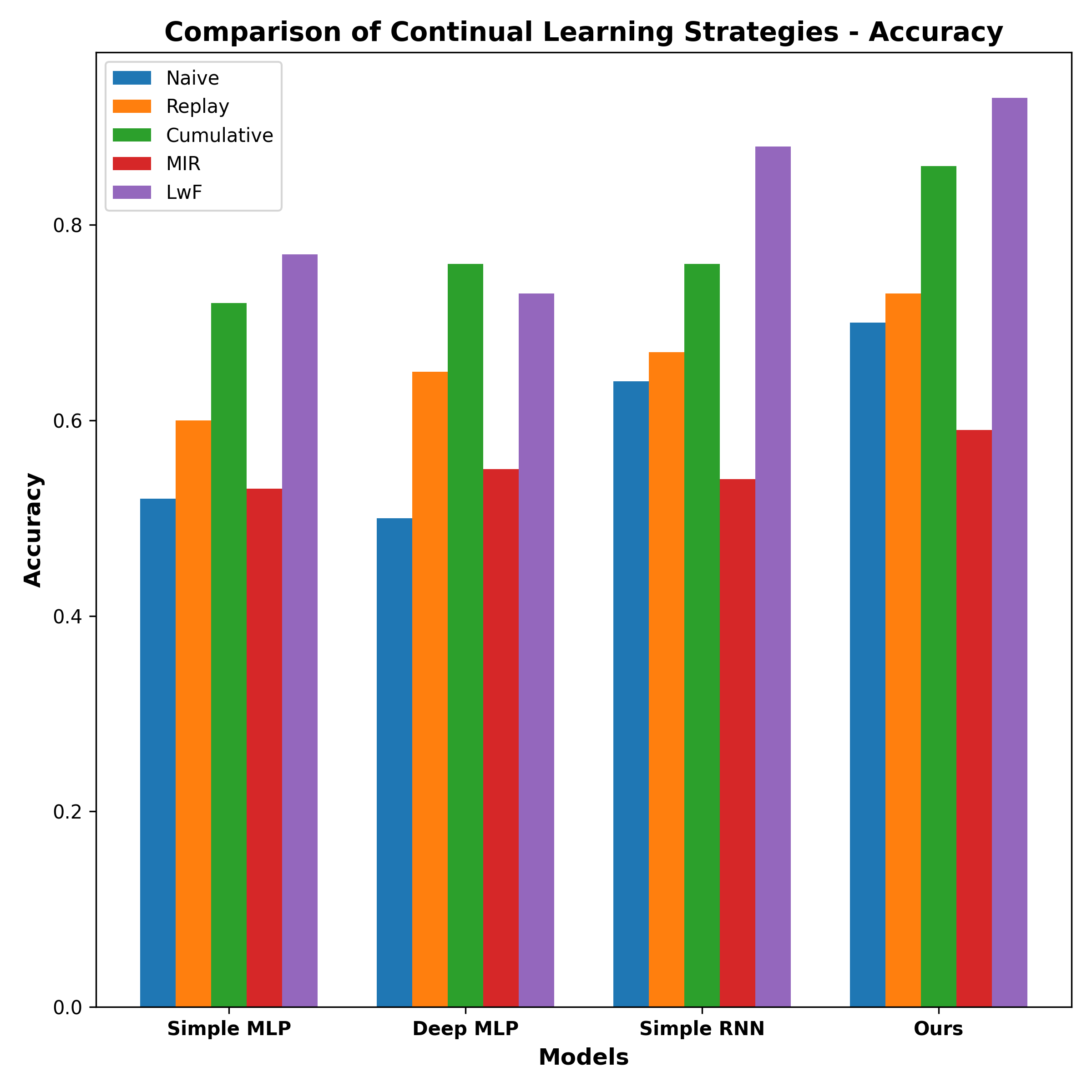} % Adjust the width as needed
\caption{Accuracy comparison.}
\label{acc}
\end{figure}

\begin{figure}[htbp]
\centering
\includegraphics[width=0.8\linewidth]{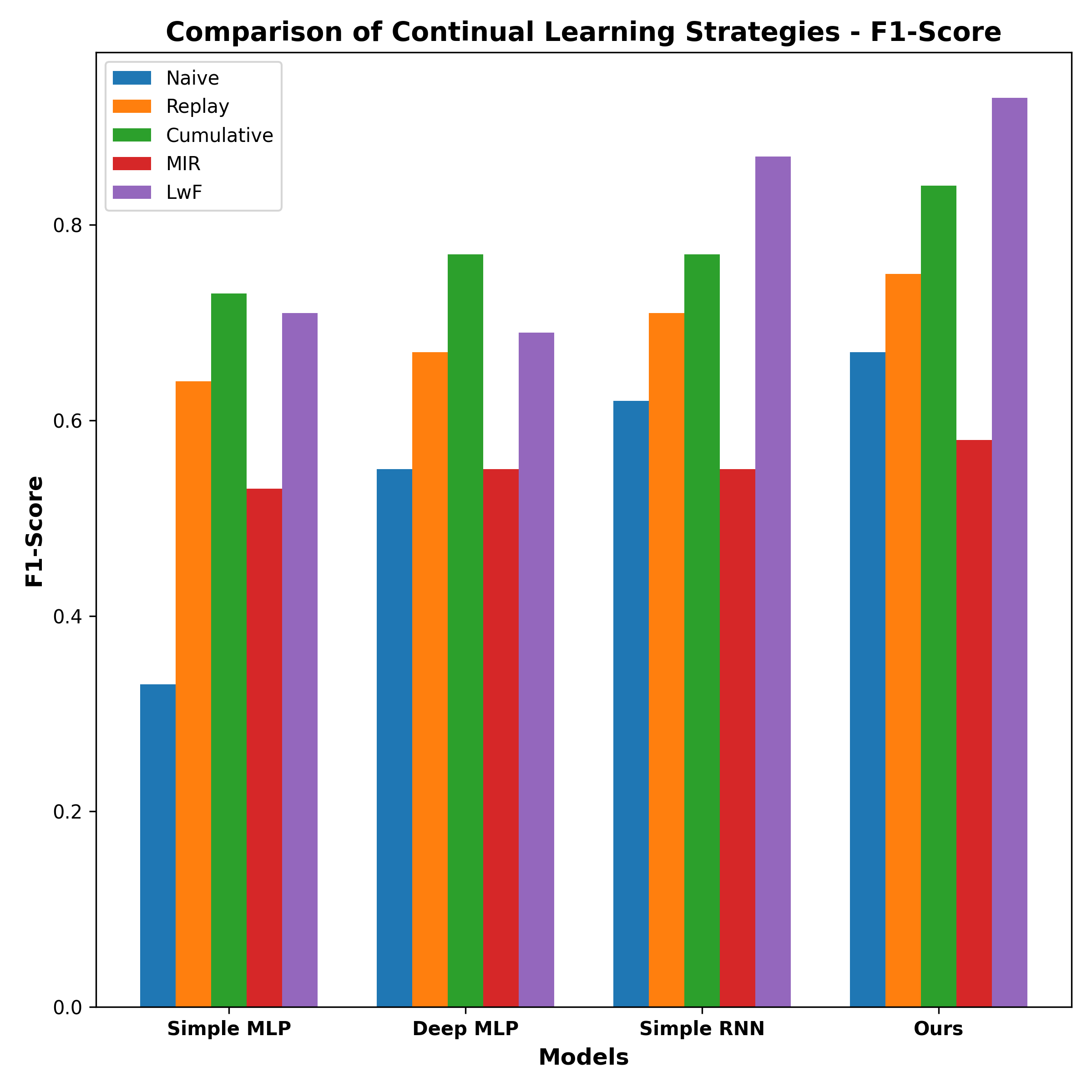} % Adjust the width as needed
\caption{F1-Score comparison.}
\label{f1}
\end{figure}

\begin{figure}[htbp]
\centering
\includegraphics[width=0.8\linewidth]{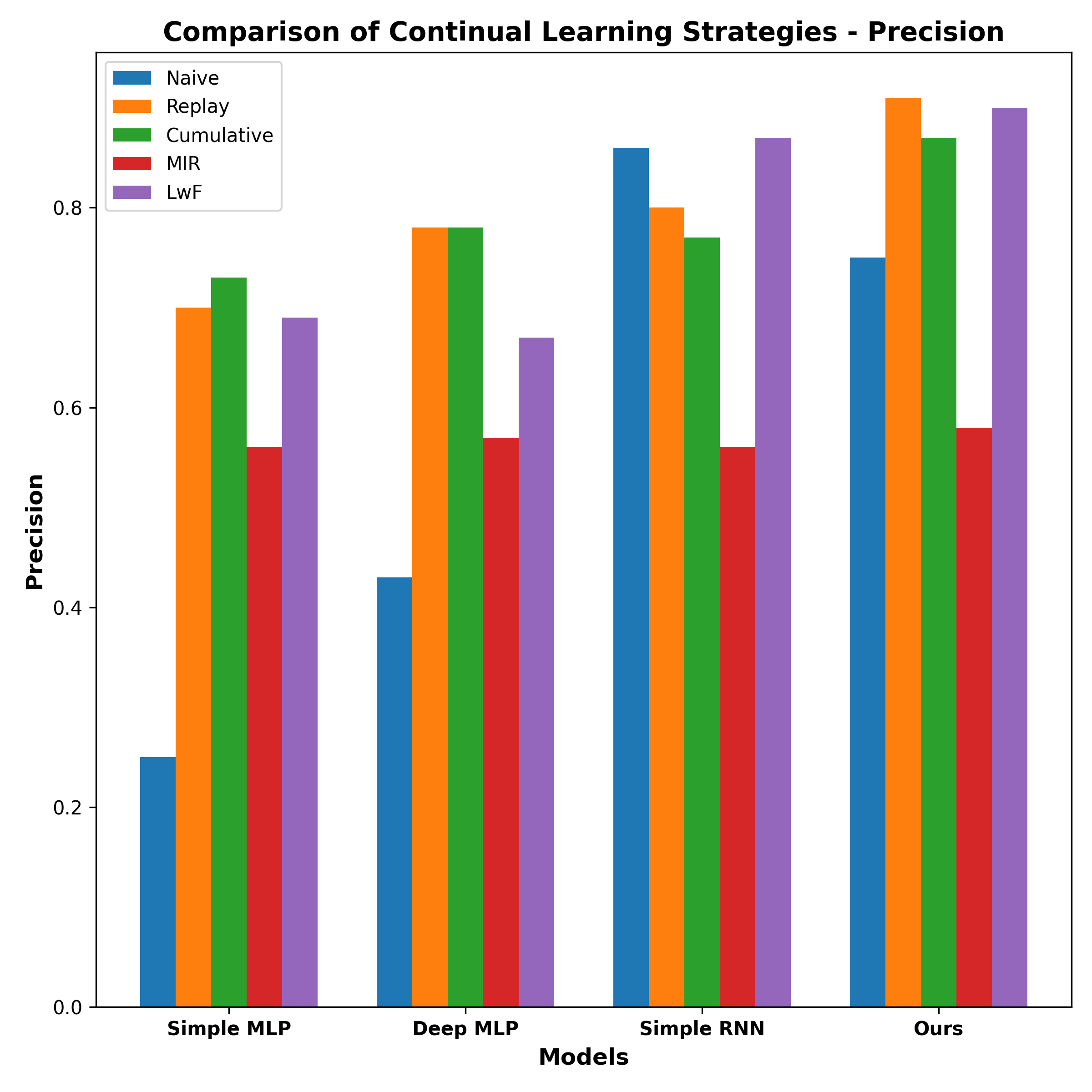} % Adjust the width as needed
\caption{Precision comparison.}
\label{prec}
\end{figure}

\begin{figure}[htbp]
\centering
\includegraphics[width=0.8\linewidth]{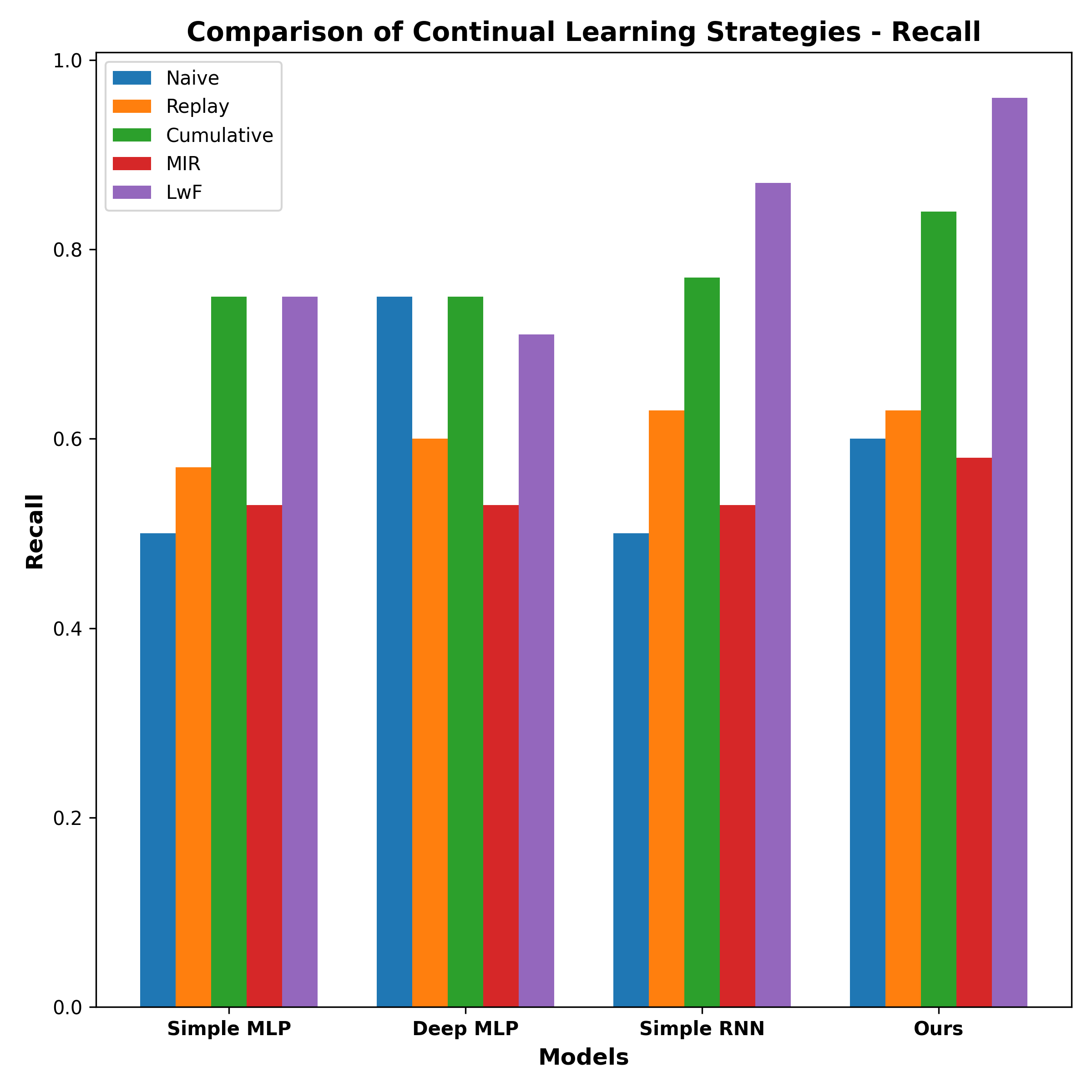} % Adjust the width as needed
\caption{Recall comparison.}
\label{rec}
\end{figure}

In Fig. \ref{acc}, Fig. \ref{f1}, Fig. \ref{prec}, Fig. \ref{rec}, through our comprehensive empirical evaluation and comparative analysis, we demonstrate the superior performance of our proposed hybrid learning paradigm and attention-based classifier model in detecting the latest phishing threats while preserving knowledge from past data distributions. Our results highlight the advantages of integrating federated and continual learning, leveraging attention mechanisms, and incorporating adaptive feature selection for robust and adaptable phishing website detection.

In summary, our proposed solution addresses the limitations of existing approaches by introducing a novel hybrid learning paradigm that combines federated and continual learning, enabling distributed nodes to collaboratively train and continually adapt a shared model to emerging phishing threats. The attention-based classifier model, tailored explicitly for web phishing detection, leverages attention mechanisms and adaptive feature selection to capture intricate patterns and distinguish between legitimate and malicious websites effectively. Through our empirical evaluation and comparative analysis, we demonstrate the efficacy and robustness of our approach, contributing to the ongoing efforts in mitigating the persistent threat of phishing attacks.

\section{Future Scope}

While our proposed approach has demonstrated promising results in addressing the challenges of robust and adaptable phishing website detection, several avenues for future research and enhancement can be explored:

\subsection{Expanded Continual Learning Strategies}

Our current work investigates several continual learning strategies, including cumulative learning, replay-based methods, and regularization-based approaches like MIR and LwF. However, the field of continual learning is rapidly evolving, and new strategies, such as meta-learning-based methods, generative replay, and hybrid approaches, can be explored to further improve the model's ability to adapt to emerging phishing threats while mitigating catastrophic forgetting.

\subsection{Multimodal Phishing Detection}

While our focus has been on leveraging textual and structural features for phishing detection, future research can explore the integration of multimodal data sources, such as visual elements (e.g., website screenshots, logos, images) and behavioral patterns (e.g., user interactions, mouse movements). By fusing these diverse modalities, our attention-based classifier model could potentially capture more sophisticated phishing patterns and improve detection accuracy.

\subsection{Adversarial Robustness}

As phishing attacks continue to evolve, adversaries may employ techniques to evade detection systems, such as adversarial examples or evasion attacks. Investigating methods to enhance the robustness of our model against such adversarial attacks is crucial for ensuring its long-term effectiveness and security.

\subsection{Interpretability and Explainability}

While our attention mechanism provides insights into the most relevant features for phishing detection, further research can be conducted to improve the interpretability and explainability of our model's decisions. This could involve developing techniques for visualizing and explaining the attention weights, as well as integrating human-in-the-loop approaches for interactive model refinement and knowledge extraction.

\subsection{Real-World Deployment and Continuous Monitoring}

To fully realize the potential of our approach, future work should focus on deploying and continuously monitoring our system in real-world scenarios. This would involve addressing practical challenges such as scalability, system integration, and data privacy concerns, as well as establishing mechanisms for continuous model updates and performance monitoring.

\subsection{Cross-Domain Transfer Learning}

Exploring the applicability of our approach to other domains beyond phishing website detection could open up new research avenues. By leveraging transfer learning techniques, our attention-based classifier model could potentially be adapted and fine-tuned for detecting other types of cyber threats or malicious activities, such as malware detection, spam filtering, or fraudulent activity identification.

By pursuing these future research directions, we can further advance the field of phishing website detection and contribute to the development of more robust, adaptable, and secure systems for protecting users and organizations against the ever-evolving landscape of cyber threats.

\section{Conclusion}

In this research, we have proposed a novel hybrid learning paradigm that seamlessly integrates federated learning and continual learning to address the challenges of robust and adaptable phishing website detection. By enabling distributed nodes to collaboratively train and continually adapt a shared model, our approach ensures timely detection of emerging phishing threats while preserving user privacy and data sovereignty.

At the core of our solution is a tailored attention-based classifier model designed explicitly for web phishing detection. This model leverages attention mechanisms to capture intricate patterns and contextual cues indicative of phishing websites, enhancing the accuracy and robustness of the detection process. Furthermore, we incorporate adaptive feature selection mechanisms to dynamically identify the most relevant features, improving the model's performance and interpretability.

Through an extensive empirical evaluation, we have demonstrated the superior performance of our proposed approach in detecting the latest phishing threats while preserving knowledge from past data distributions. By comparing our approach with traditional machine learning techniques and state-of-the-art methods across various continual learning strategies, model architectures, and datasets, we have shown that our model combined with the Learning without Forgetting (LwF) continual learning strategy yields the most robust results.

By addressing the limitations of existing approaches and offering a comprehensive solution for robust and adaptable phishing detection, our work contributes significantly to the ongoing efforts in mitigating the persistent threat of phishing attacks, ultimately enhancing online security and protecting users from falling victim to these deceptive attacks.

In conclusion, our proposed hybrid learning paradigm and attention-based classifier model represent a significant step forward in the battle against phishing attacks, offering a robust and adaptable solution that can effectively detect and mitigate these persistent threats, safeguarding online users and promoting a more secure digital ecosystem.

\nocite{*}
\bibliographystyle{plain}
\bibliography{ref} 

\begin{thebibliography}{10}

\bibitem{15}
Saleem~Raja Abdul~Samad, Sundarvadivazhagan Balasubaramanian, Amna~Salim Al-Kaabi, Bhisham Sharma, Subrata Chowdhury, Abolfazl Mehbodniya, Julian~L Webber, and Ali Bostani.
\newblock Analysis of the performance impact of fine-tuned machine learning model for phishing url detection.
\newblock {\em Electronics}, 12(7):1642, 2023.

\bibitem{10}
R~Alazaidah, A~Al-Shaikh, MR~AL-Mousa, H~Khafajah, G~Samara, M~Alzyoud, N~Al-Shanableh, and S~Almatarneh.
\newblock Website phishing detection using machine learning techniques.
\newblock {\em Journal of Statistics Applications \& Probability}, 13(1):119--129, 2024.

\bibitem{25}
Rahaf Aljundi et~al.
\newblock Online continual learning with maximal interfered retrieval.
\newblock In {\em Advances in Neural Information Processing Systems}, volume~32, 2019.

\bibitem{8}
Shouq Alnemari and Majid Alshammari.
\newblock Detecting phishing domains using machine learning.
\newblock {\em Applied Sciences}, 13(8):4649, 2023.

\bibitem{18}
Ahmet~Selman Bozkir, Firat~Coskun Dalgic, and Murat Aydos.
\newblock Grambeddings: a new neural network for url based identification of phishing web pages through n-gram embeddings.
\newblock {\em Computers \& Security}, 124:102964, 2023.

\bibitem{12}
Sumitra Das~Guptta, Khandaker~Tayef Shahriar, Hamed Alqahtani, Dheyaaldin Alsalman, and Iqbal~H Sarker.
\newblock Modeling hybrid feature-based phishing websites detection using machine learning techniques.
\newblock {\em Annals of Data Science}, 11(1):217--242, 2024.

\bibitem{3}
Ashit~Kumar Dutta.
\newblock Detecting phishing websites using machine learning technique.
\newblock {\em PloS one}, 16(10):e0258361, 2021.

\bibitem{11}
Asif Ejaz, Adnan~Noor Mian, and Sanaullah Manzoor.
\newblock Life-long phishing attack detection using continual learning.
\newblock {\em Scientific Reports}, 13(1):11488, 2023.

\bibitem{21}
Abdelhakim Hannousse and Salima Yahiouche.
\newblock Towards benchmark datasets for machine learning based website phishing detection: An experimental study.
\newblock {\em Engineering Applications of Artificial Intelligence}, 104:104347, 2021.

\bibitem{10.1007/978-3-031-59164-8_28}
M.~Jesher Joshua, V.~Ragav, and S.~P.~Syed Ibrahim.
\newblock Advanced knowledge extraction of physical design drawings, translation and conversion to cad formats using deep learning.
\newblock In Pichai Janmanee, Saichol Chujuarjeen, Suthep Butdee, Phatchani Srikhumsuk, Andre D.~L. Batako, Anna Burduk, and M.~Anthony Xavior, editors, {\em Advanced in Creative Technology- added Value Innovations in Engineering, Materials and Manufacturing}, pages 343--356, Cham, 2024. Springer Nature Switzerland.

\bibitem{16}
Luka Jovanovic, Dijana Jovanovic, Milos Antonijevic, Bosko Nikolic, Nebojsa Bacanin, Miodrag Zivkovic, and Ivana Strumberger.
\newblock Improving phishing website detection using a hybrid two-level framework for feature selection and xgboost tuning.
\newblock {\em Journal of Web Engineering}, 22(3):543--574, 2023.

\bibitem{20}
Parvathapuram~Pavan Kumar, T~Jaya, and V~Rajendran.
\newblock Si-bba--a novel phishing website detection based on swarm intelligence with deep learning.
\newblock {\em Materials Today: Proceedings}, 80:3129--3139, 2023.

\bibitem{23}
Zhizhong Li and Derek Hoiem.
\newblock Learning without forgetting.
\newblock {\em IEEE Transactions on Pattern Analysis and Machine Intelligence}, 40(12):2935--2947, 2017.

\bibitem{7}
Antonio Maci, Alessandro Santorsola, Antonio Coscia, and Andrea Iannacone.
\newblock Unbalanced web phishing classification through deep reinforcement learning.
\newblock {\em Computers}, 12(6):118, 2023.

\bibitem{4}
Naya Nagy, Malak Aljabri, Afrah Shaahid, Amnah~Albin Ahmed, Fatima Alnasser, Linda Almakramy, Manar Alhadab, and Shahad Alfaddagh.
\newblock Phishing urls detection using sequential and parallel ml techniques: comparative analysis.
\newblock {\em Sensors}, 23(7):3467, 2023.

\bibitem{6}
Naya Nagy, Malak Aljabri, Afrah Shaahid, Amnah~Albin Ahmed, Fatima Alnasser, Linda Almakramy, Manar Alhadab, and Shahad Alfaddagh.
\newblock Phishing urls detection using sequential and parallel ml techniques: comparative analysis.
\newblock {\em Sensors}, 23(7):3467, 2023.

\bibitem{19}
Alper Ozcan, Cagatay Catal, Emrah Donmez, and Behcet Senturk.
\newblock A hybrid dnn--lstm model for detecting phishing urls.
\newblock {\em Neural Computing and Applications}, pages 1--17, 2023.

\bibitem{PANDIYARAJU2024106436}
V.~Pandiyaraju, Sannasi Ganapathy, A.M. {Senthil Kumar}, M.~{Jesher Joshua}, V.~Ragav, S.~{Sree Dananjay}, and A.~Kannan.
\newblock A new clinical diagnosis system for detecting brain tumor using integrated resnet stacking with xgboost.
\newblock {\em Biomedical Signal Processing and Control}, 96:106436, 2024.

\bibitem{17}
Manoj~Kumar Prabakaran, Parvathy Meenakshi~Sundaram, and Abinaya~Devi Chandrasekar.
\newblock An enhanced deep learning-based phishing detection mechanism to effectively identify malicious urls using variational autoencoders.
\newblock {\em IET Information Security}, 17(3):423--440, 2023.

\bibitem{13}
Saba~Hussein Rashid and Wisam~Dawood Abdullah.
\newblock Enhanced website phishing detection based on the cyber kill chain and cloud computing.
\newblock {\em Indonesian Journal of Electrical Engineering and Computer Science}, 32(1):517--529, 2023.

\bibitem{14}
Nitin~N Sakhare, Jyoti~L Bangare, Radhika~G Purandare, Disha~S Wankhede, and Pooja Dehankar.
\newblock Phishing website detection using advanced machine learning techniques.
\newblock {\em International Journal of Intelligent Systems and Applications in Engineering}, 12(12s):329--346, 2024.

\bibitem{5}
Said Salloum, Tarek Gaber, Sunil Vadera, and Khaled Shaalan.
\newblock Phishing website detection from urls using classical machine learning ann model.
\newblock In {\em International Conference on Security and Privacy in Communication Systems}, pages 509--523. Springer, 2021.

\bibitem{10157704}
Abhishek Sebastian, Pragna R, Madhan~Kumar S, Jesher~Joshua M, Sarath Prathap, and Brintha~Therese A.
\newblock Proximity-based access control with ble communication using path loss model and mlp prediction.
\newblock In {\em 2023 2nd International Conference on Vision Towards Emerging Trends in Communication and Networking Technologies (ViTECoN)}, pages 1--6, 2023.

\bibitem{9}
Sanjeev Shukla, Manoj Misra, and Gaurav Varshney.
\newblock Http header based phishing attack detection using machine learning.
\newblock {\em Transactions on Emerging Telecommunications Technologies}, 35(1):e4872, 2024.

\bibitem{2}
Fu~Song, Yusi Lei, Sen Chen, Lingling Fan, and Yang Liu.
\newblock Advanced evasion attacks and mitigations on practical ml-based phishing website classifiers.
\newblock {\em International Journal of Intelligent Systems}, 36(9):5210--5240, 2021.

\bibitem{24}
K.~Thórisson et~al.
\newblock Cumulative learning.
\newblock In {\em International Conference on Artificial General Intelligence}, volume~7, pages 198--208. Springer, 2019.

\bibitem{v2024comprehensive}
Ragav V, Jesher~Joshua M, and Syed Ibrahim~S P.
\newblock Comprehensive autonomous vehicle optimal routing with dynamic heuristics, 2024.

\bibitem{22}
Grega Vrbančič.
\newblock Phishing websites dataset, 2020.

\bibitem{1}
Ammara Zamir, Hikmat~Ullah Khan, Tassawar Iqbal, Nazish Yousaf, Farah Aslam, Almas Anjum, and Maryam Hamdani.
\newblock Phishing website detection using diverse machine learning algorithms.
\newblock {\em The Electronic Library}, 38(1):65--80, 2020.

\end{thebibliography}
\end{document}